\documentclass[10pt,twocolumn,letterpaper]{article}

\usepackage[pagenumbers]{cvpr}

\usepackage{multicol}
\usepackage{multirow}
\usepackage{wrapfig}
\usepackage[dvipsnames]{xcolor}

\definecolor{cvprblue}{rgb}{0.21,0.49,0.74}
\usepackage[pagebackref,breaklinks,colorlinks,citecolor=cvprblue]{hyperref}

\def\paperID{}
\def\confName{}
\def\confYear{}

\title{ARM: Appearance Reconstruction Model for Relightable 3D Generation}

\author{
    Xiang Feng$^{1,2\ast}$\ \ \ \ \ 
    Chang Yu$^{3\ast}$\ \ \ \ \ \ 
    Zoubin Bi$^{2\ast}$\ \ \ \ \ 
    Yintong Shang$^{1}$\ \ \ \ \ 
    Feng Gao$^{4}$
    \\
    Hongzhi Wu$^{2}$\ \ \ \ \ 
    Kun Zhou$^{2}$\ \ \ \ \ 
    Chenfanfu Jiang$^{3}$\ \ \ \ \ 
    Yin Yang$^{1}$
    \\
    $^{1}$University of Utah\ \ \ \ \ 
    $^{2}$Zhejiang University\ \ \ \ \ 
    $^{3}$UCLA\ \ \ \ \ 
    $^{4}$Amazon
    \\
    \url{https://arm-aigc.github.io}
}

\usepackage{ulem}
\usepackage{lipsum}

\newcommand{\secref}[1]{Sec.~\ref{#1}}
\newcommand{\figref}[1]{Fig.~\ref{#1}}
\newcommand{\tabref}[1]{Tab.~\ref{#1}}

\newcommand{\eqnref}[1]{Eq.~\ref{#1}}

\newcommand\blfootnote[1]{
\begingroup
\renewcommand\thefootnote{}\footnote{#1}
\addtocounter{footnote}{-1}
\endgroup
}

\begin{document}

\twocolumn[{
\renewcommand\twocolumn[1][]{#1}
\maketitle
\begin{center}
    \centering
    \vspace{-10pt}
    \begin{minipage}[t]{\textwidth}
        \centering
        \includegraphics[width=\textwidth]{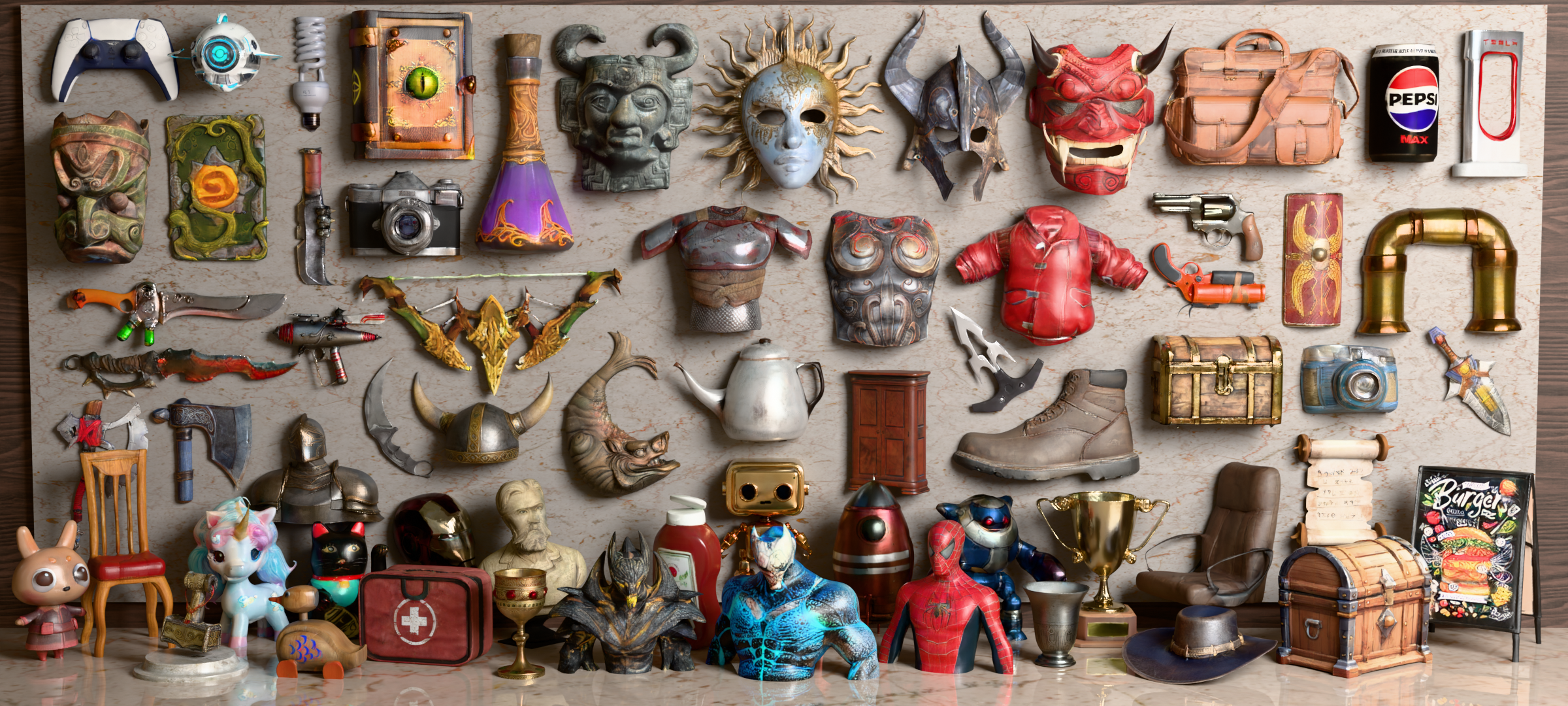}
    \end{minipage}
    \captionsetup{type=figure}
    \vspace{-10pt}
    \captionof{figure}{\textbf{ARM generates high-quality, relightable 3D content from a single image input.} This figure presents sample results generated from different input images, demonstrating ARM's ability to reconstruct a variety of objects with spatially-varying appearance. Please refer to our supplementary video for results under dynamic view and lighting.}
    \label{fig:teaser}
\end{center}
}]

\begin{abstract}
\blfootnote{$\ast$ indicates equal contributions.} 
Recent image-to-3D reconstruction models have greatly advanced geometry generation, but they still struggle to faithfully generate realistic appearance. To address this, we introduce ARM, a novel method that reconstructs high-quality 3D meshes and realistic appearance from sparse-view images. The core of ARM lies in decoupling geometry from appearance, processing appearance within the UV texture space. Unlike previous methods, ARM improves texture quality by explicitly back-projecting measurements onto the texture map and processing them in a UV space module with a global receptive field. To resolve ambiguities between material and illumination in input images, ARM introduces a material prior that encodes semantic appearance information, enhancing the robustness of appearance decomposition. Trained on just 8 H100 GPUs, ARM outperforms existing methods both quantitatively and qualitatively.
\end{abstract}

\section{Introduction}
Obtaining high-quality 3D mesh models with realistic appearance from 2D images has become a critical task in computer vision and computer graphics, with applications spanning the metaverse, gaming, and e-commerce. Conventionly, 3D models with appearance are hand-crafted by skilled artists using specialized modeling software—a highly time-consuming process that can take hours or even days. Alternatively, 3D models can be reconstructed from multi-view input images using optimization-based approaches~\cite{wang2021neus,Munkberg_2022_CVPR, mildenhall2021nerf,kerbl20233d}, typically requiring over a hundred images from different viewpoints. Photometric devices are often needed to accurately capture high-quality appearance~\cite{meka2020deep,Kang:2019:JOINT,nam2018practical,saito2024relightable,Ren_2022_CVPR}.

Recent advances in 3D generation and reconstruction models have brought new insights to the field, leading to several dominant approaches. One line of research~\cite{poole2022dreamfusion,chen2023fantasia3d,lin2023magic3d,shi2023mvdream,wang2024prolificdreamer,liu2023zero1to3} leverages priors from pretrained 2D diffusion models to distill 3D shapes, typically by generating images from multiple viewpoints with a 2D generative model, followed by per-scene optimization. However, this distillation process is time-consuming, limiting its practicality for real-world applications. Another line~\cite{hong2023lrm,xu2024instantmesh,wei2024meshlrm,tang2024lgm,tochilkin2024triposr} trains a feed-forward neural network directly on large-scale 3D datasets~\cite{deitke2023objaverse,deitke2024objaverse} to learn 3D priors from single-view or sparse-view input images, an approach our method also adopts. This strategy can improve consistency across multi-views and achieves faster inference compared to time-intensive distillation techniques. However, existing methods in this area still struggle with several limitations: the reconstructed textures often appear blurred and lack fine details, leading to overall low-quality results. Moreover, most current methods represent object appearance only through per-vertex colors, which is even simpler than Lambertian shading and includes baked-in reflections, lighting, and shadows. This simplified shading model fails to capture realistic view-dependent and lighting-dependent effects, making the generated assets unsuitable for downstream applications such as gaming or metaverse, where dynamic lighting and viewpoint changes are essential for realism. 

In this work, we introduce a framework called ARM for reconstructing high-quality 3D meshes with fine-detailed textures and realistic appearance. ARM builds on Large Reconstruction Models (LRMs)~\cite{hong2023lrm}, using triplanes as its 3D representation. While LRMs offer strong geometry capabilities, we observed that reconstructed textures often appear overly blurred due to the limited resolution of triplanes and the relatively small decoding MLPs. ARM’s core innovation lies in decoupling geometry generation from appearance modeling by processing appearance directly within the UV space. Unlike previous methods that decode color from learned triplanes using MLPs, ARM enhances texture quality by explicitly back-projecting multi-view measurements onto the texture map and processing them with a UV-space module featuring a global receptive field. The UV texture space offers advantages over the  triplane space by directly representing color variations on the object surface. ARM also introduces an approach to address the material and illumination ambiguities present in sparse-view input images—--a fundamentally ill-posed problem. Previous methods~\cite{siddiqui2024meta} often attempt to tackle this issue using a rendering loss; however, these inverse rendering approaches have been shown to struggle with sparse inputs and may even fail with dense multi-view data. In contrast, ARM incorporates a material prior that encodes semantic appearance information and directly fits to ground-truth materials, enhancing the robustness of appearance decomposition. Experimental results show that ARM outperforms recent image-to-3D methods both qualitatively and quantitatively, demonstrating its capability to generate versatile objects with realistic appearance.

\section{Related work}

\subsection{3D generation with 2D diffusion priors}

The emergence of 2D diffusion models~\cite{ho2020denoising, saharia2022photorealistic, rombach2022high} has driven significant advancements in 3D generation. DreamFusion~\cite{poole2022dreamfusion} was the first to use SDS~\cite{wang2023score} loss to iteratively distill 3D representations from a 2D diffusion model, inspiring numerous follow-up works~\cite{shi2023mvdream, lin2023magic3d, seo2023let, wang2023imagedream, metzer2023latent, chen2023fantasia3d, lorraine2023att3d, long2024wonder3d, wu2024unique3d, sun2023dreamcraft3d}. These methods typically convert multi-view images generated by diffusion models into 3D representations such as NeRF~\cite{mildenhall2021nerf}, NeuS~\cite{wang2021neus}, and 3DGS~\cite{kerbl20233d}, using SDS-like losses and optimization-based techniques. Subsequent work has focused on enhancing cross-view consistency and generalization. For example, Zero-1-to-3~\cite{liu2023zero1to3} leverages diffusion models to create consistent multi-view images for 3D reconstruction, while other methods~\cite{liu2023syncdreamer, qian2023magic123, shi2023zero123++, liu2023text, lin2024consistent123, chan2023generative, ye2024consistent, hu2024mvd} further improve consistency through conditioning and distribution modeling. Wang et al.~\cite{wang2024prolificdreamer} proposed Variational Score Distillation (VSD) to address over-saturation in SDS and improve texture diversity. Recently, Voleti et al.~\cite{voleti2025sv3d} introduced an image-to-video diffusion model for enhanced generalization and multi-view consistency in novel view synthesis. Despite these advancements, per-shape optimization methods still face key limitations, with long runtimes as the main bottleneck to broader application. Issues with Janus inconsistencies and over-saturation also remain.

\begin{figure*}
    \centering
    \includegraphics[width=\textwidth]{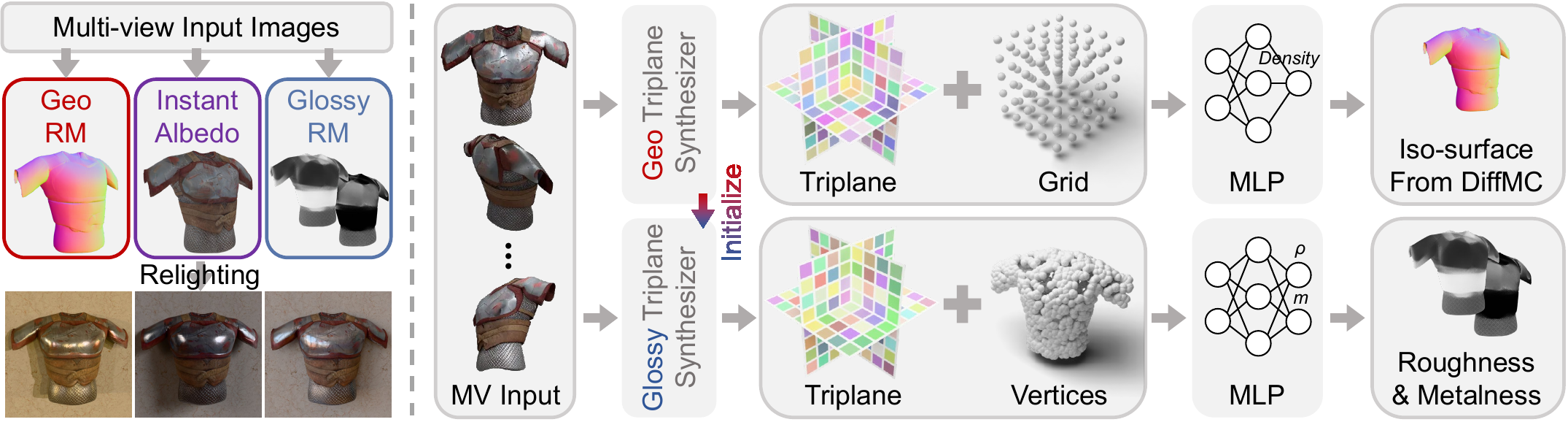}
    \caption{\textbf{Overview of our pipeline.} \textbf{(left)} Starting from sparse-view input images generated by a diffusion model~\cite{shi2023zero123++}, ARM separates shape and appearance generation into two stages. In the geometry stage, ARM uses GeoRM to predict a 3D shape from the input images. In the appearance stage, ARM employs InstantAlbedo and GlossyRM to reconstruct PBR maps, enabling realistic relighting under varied lighting conditions. \textbf{(right)} Both GeoRM and GlossyRM share the same architecture, consisting of a triplane synthesizer and a decoding MLP. GeoRM is trained to predict density and extracts an iso-surface from the density grid with DiffMC~\cite{wei2023neumanifold}, while GlossyRM is trained to predict roughness and metalness. GlossyRM is trained after GeoRM and initializes with the weights of GeoRM at the start of training.}
    \label{fig:overview}
\end{figure*}

\subsection{3D generation with 3D feed-forward models}

With the availability of high-quality, large-scale annotated 3D datasets~\cite{deitke2023objaverse,deitke2024objaverse}, recent works have focused on generating 3D objects using feed-forward models~\cite{hong2023lrm, jun2023shap, xie2024latte3d, ren2024xcube}, achieving significantly faster speeds than score-distillation methods. For example, One-2-3-45~\cite{liu2024one} uses the generalizable NeRF method for 3D reconstruction, while One-2-3-45++~\cite{liu2023one2345++} improves quality with a two-stage, coarse-to-fine 3D diffusion model similar to LAS-Diffusion~\cite{zheng2023locally}, though high-quality textures remain a challenge. The LRM model~\cite{hong2023lrm} and its extensions~\cite{li2023instant3d, xu2023dmv3d, wang2023pf, wang2024crm, xu2024instantmesh, wei2024meshlrm, tochilkin2024triposr} enhance generation quality through transformer-based architectures and triplane representation~\cite{chan2022efficient}. While triplanes provide an effective hybrid representation, other works~\cite{zou2024triplane, xu2024grm, tang2024lgm, zhang2025gs} explore 3DGS~\cite{kerbl20233d} for potentially faster generation. Recently, MeshFormer~\cite{liu2024meshformer} replaced triplanes with 3D sparse voxels to represent fine-grained shapes explicitly. Our method also adopts a feed-forward approach for efficient generation. Using LRM as the backbone, we address a key limitation: relying solely on triplanes restricts texture detail due to resolution constraints, and scaling up the network is impractical due to memory demands. Instead, we directly learn textures in UV space, significantly enhancing texture quality.

\subsection{Material generation and decomposition}

Estimating surface material properties remains a longstanding challenge in 3D reconstruction and generation, as joint optimization of unknown lighting and appearance makes this problem inherently ill-posed. Recent advances in multi-view reconstruction, including NeRF-based methods~\cite{zhang2021nerfactor, srinivasan2021nerv, boss2021nerd, zhang2021physg, boss2022samurai, munkberg2022extracting, hasselgren2022shape, engelhardt2024shinobi, zeng2023nrhints} and 3DGS-based approaches~\cite{gao2023relightable, bi2024gs, liang2024gs,jiang2024gaussianshader}, show promise for material estimation but still require dense multi-view input. SDS-based text-to-3D pipelines~\cite{chen2023fantasia3d, xu2023matlaber, liu2025unidream} offer a solution but remain time-intensive due to SDS optimization. Make-it-Real~\cite{fang2024make} uses LLMs to identify object semantics and retrieve materials from a library, while SF3D~\cite{boss2024sf3d} adds UV unwrapping with material prediction in LRM. Other methods~\cite{vainer2025collaborative, youwang2024paint, zhang2024dreammat,zeng2024dilightnet} apply diffusion models for appearance generation on existing geometries with text prompts. Recently, CLAY~\cite{zhang2024clay} and 3DTopia-XL~\cite{chen20243dtopia} leverage Diffusion Transformer models (DiT)~\cite{peebles2023scalable} for 3D asset generation with appearance. Our method differs by being the first to perform material decomposition entirely in UV space, conditioned directly on in-the-wild image input. Since our network inherently learns texture-level priors through back-projection in the first step, it mitigates the inconsistent artifacts caused by generating multi-view appearance parameters first and then back-projecting them onto textures. Additionally, by introducing a material prior in texture-level appearance decomposition, our method more accurately disentangles illumination from materials, achieving greater fidelity than previous approaches.

\section{Preliminaries}
We model the spatially varying, view-dependent, and lighting-dependent appearance of objects by an Spatially Varying Bi-directional Reflectance Distribution Function (SVBRDF), where each surface point’s reflectance is modeled by a microfacet BRDF parameterized by its diffuse albedo $\mathbf{c}_d$, roughness $\rho$, and metalness $m$. The BRDF $f_{\rm r}$ consists of a diffuse component and a glossy component:
\begin{equation}
    f_{\rm r}(\mathbf{l}, \mathbf{v})=(1-m)\frac{\mathbf{c}_d}{\pi}+\frac{DFG}{4(\mathbf{n}\cdot\mathbf{l})(\mathbf{n}\cdot\mathbf{v})}
    \label{eq:brdf}
\end{equation}
where $\mathbf{l}$ and $\mathbf{v}$ are the directions of the incoming light and the view, $\mathbf{n}$ is the surface normal, and $D$, $F$, and $G$ represent the microfacet normal distribution, Fresnel, and geometry terms, respectively. These terms are determined by $\rho$, $m$, and $\mathbf{c}_d$, with detailed expressions provided in the supplementary material. In summary, ARM reconstructs spatially varying diffuse, roughness, and metalness maps—also known as Physically-Based Rendering (PBR) maps~\cite{burley2012physically}—to efficiently capture diverse material appearances.

\section{Overview}
We propose ARM, a framework for simultaneously reconstructing high-quality 3D meshes and PBR texture maps, as illustred in \figref{fig:overview}. Starting from sparse-view images generated from a single view using a diffusion model~\cite{shi2023zero123++}, ARM separates shape and appearance generation into two stages. In the geometry stage, ARM employs \textit{GeoRM} (\secref{sec:decouple_shape_app}) to predict a 3D shape from input images, which is then unwrapped into atlas charts for latter processing in the UV texture space. 
\begin{figure*}
    \centering
    \includegraphics[width=\textwidth]{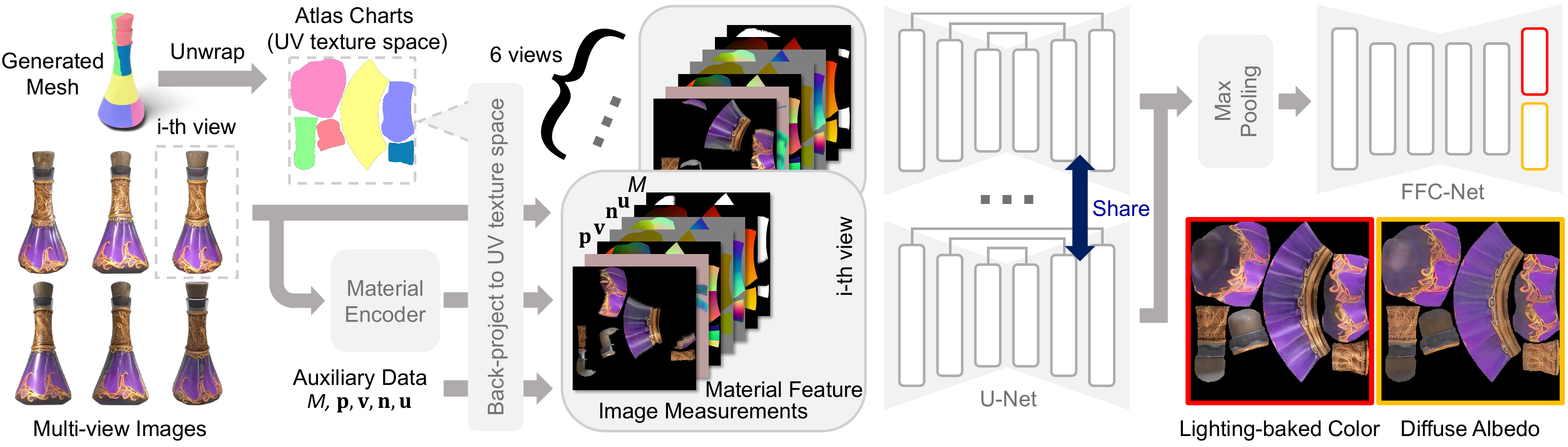}
    \caption{\textbf{Overview of InstantAlbedo.} InstantAlbedo operates in the texture UV space. This process begins by converting all necessary data to UV texture space. Given the unwrapped mesh from GeoRM, we back-project images, material encodings, and auxiliary data into UV texture space, resulting in six sets of inputs corresponding to the six input views. InstantAlbedo then processes these maps using a U-Net and an inpainting-specific FFC-Net to predict both the lighting-baked color and the decomposed diffuse albedo UV textures. }
    \vspace{-10pt}
    \label{fig:instantalbedo}
\end{figure*}
In the appearance stage, ARM introduces two specialized models: \textit{InstantAlbedo} (\secref{sec:instant_albedo}) and \textit{GlossyRM} (\secref{sec:decouple_shape_app}). Both models take input as aforementioned sparse-view input images and the predicted 3D shape from GeoRM. GlossyRM outputs per-vertex roughness $\rho$ and metalness $m$, capturing the glossy component of the appearance. InstantAlbedo, in contrast, operates in the UV texture space, producing two color maps: one with baked-in lighting and another with diffuse albedo, representing the diffuse component of appearance. To enhance the accuracy and robustness of albedo decomposition, ARM also incorporates a material prior (\secref{sec:material_prior}) into InstantAlbedo to encode semantic appearance information from the multi-view images. The final material is stored as PBR texture maps, enabling realistic relighting under novel lighting.

\section{Method}

In this section, we present the design of ARM and outline the key insights and motivations behind our approach.

\subsection{Decoupling shape and appearance}
\label{sec:decouple_shape_app}
To produce high-quality 3D meshes with high-fidelity reflective properties, ARM introduces two distinct LRM-based models: GeoRM and GlossyRM. Both models share the same architecture, including a transformer-based triplane synthesizer and a decoding MLP. The triplane, as an internal 3D representation, is generated by the synthesizer from sparse-view input images. Any queried 3D point is then projected onto the triplane to retrieve its corresponding feature, which is decoded by the MLP to produce the desired output. By decoupling shape and appearance, ARM designs the two models to serve different purposes, each trained with its own objective function.

GeoRM is trained to predict the density, with the iso-surface extracted from a $256^3$ density grid using differentiable marching cubes~\cite{wei2023neumanifold}. Optimized solely with geometry-related losses (mask, depth, and normal supervision), GeoRM’s purpose is exclusively to generate geometry, and its weights are frozen once trained.

GlossyRM, trained after GeoRM, uses the 3D mesh generated by GeoRM to query its own learned triplane and decode per-vertex roughness $\rho$ and metalness $m$. Its training is focused solely on losses related to these parameters.

The core idea here is to decouple geometry from appearance, enabling (1) efficient processing in texture space, since unwrapping geometry into texture space is time-consuming and impractical during training, and (2) increased network capacity for both GeoRM and GlossyRM, with each model focused on a specific, smaller task. A naive approach would be to train a single LRM to predict all desired outputs---density, vertex color, and material parameters---but our pilot study showed that as the number of prediction targets increases, output quality degrades significantly, with results becoming blurred. This issue is particularly pronounced with material predictions, as these parameters are more challenging to infer than colors directly visible in the input images, thereby increasing learning complexity.

Furthermore, we observe that the triplane resolution is directly correlated with shape quality and the presence of voxel artifacts, which can be mitigated by increasing the resolution. To address this, we introduce a super-resolution module that raises triplane resolution to $256\times256$, significantly enhancing shape reconstruction quality. By decoupling geometry and appearance, the memory footprint of both models is reduced, making this resolution increase feasible. Further details on the architecture of GeoRM and GlossyRM can be found in our supplementary material.

\subsection{Appearance decomposition in texture space}
\label{sec:instant_albedo}

The key insight of ARM is to reconstruct fine-detailed textures in UV texture space, which directly aligns with the spatial variations on the object surface, rather than relying on the triplane representation used in previous LRMs. There are two main reasons for moving away from triplanes: First, texture details are limited by the triplane resolution, which is constrained by memory usage of large transformer. Increasing triplane resolution to capture finer details quadratically
increases the transformer complexity, reducing practicality for high-detail reconstructions. Second, triplanes function similarly to volumetric representations, where querying with decoding MLPs often results in blurriness. This is because triplanes store 3D information across three planes, but the spatial variation in these planes does not directly correspond to the texture variation on the object surface, leading to interpolation mismatches and degraded texture quality.

To overcome these limitations, ARM employs InstantAlbedo, a network specifically designed to operate in UV texture space, as illustrated in \figref{fig:instantalbedo}. By working directly in texture space, InstantAlbedo captures fine surface details without the resolution and memory constraints associated with triplanes.

The process begins by converting all necessary data to UV texture space. Given the 3D mesh generated by GeoRM, we unwrap it into atlas charts in the UV texture, where each texel corresponds to a specific surface point. Since unwrapping is time-consuming and impractical to perform during training, we synthesize a pre-unwrapped training dataset offline, which we detail later. Next, with multi-view input images generated at $6$ known camera poses, we directly back-project these multi-view images onto the UV texture. Along with the images, we also back-project auxiliary data, including masks $M$, positions $\mathbf{p}$, texture coordinates $\mathbf{u}$, view directions $\mathbf{v}$, and surface normals $\mathbf{n}$. These inputs provide valid contextual information for material decomposition according to \eqnref{eq:brdf}, effectively reducing the learning complexity for the network. The back-projection process generates six sets of input maps in UV texture space, corresponding to the six views.

InstantAlbedo takes above six sets of maps as input and outputs two color maps: one with baked-in lighting and another with the decomposed diffuse albedo. In the first step, a U-Net is employed to extract per-view features from each set, yielding six feature maps. Since each view captures only a partial region of the object’s surface, it is necessary to fuse information from multiple views to form a complete texture. Inspired by PointNet~\cite{qi2016pointnet}, we use max-pooling to aggregate information across the six feature maps. However, with only six views, some regions of the object surface remain unobserved. To address this, we incorporate a FFCNet~\cite{chi2020fast,suvorov2022lama} with a global receptive field to extract information from other areas in the texture, inpaint unseen regions, and refine the fused result. As shown in \figref{fig:visual_abl}, this design significantly improves the completeness of the reconstructed texture. Please note that InstantAlbedo focuses solely on albedo decomposition rather than predicting the entire material (including roughness and metalness) in a unified manner, as we found that this approach led to inaccuracies in decomposed roughness and metalness. Additional details are provided in the supplementary material.

\subsection{Appearance encoding with a material prior}
\label{sec:material_prior}

ARM tackles the inherent ambiguity between material and illumination in sparse-view input, another challenging and inherently ill-posed problem. With fewer than ten input images, attempting to use inverse rendering with a simple rendering loss often results in flawed decompositions for glossy materials, where lighting effects may bake into the diffuse albedo to perfectly match the input images. To address this, ARM directly fits to ground-truth material instead of relying on a rendering loss to separate material and lighting. However, this approach is still challenging because any lighting in the input images can be hard for the network to ignore, often leaving traces of illumination in the final albedo output.

To improve robustness, ARM incorporates a material prior—--a material-aware image encoder—--into the back-projection process of InstantAlbedo, as illustrated in \figref{fig:instantalbedo}. By transforming multi-view input images into semantic, material-aware feature maps, InstantAlbedo back-projects these encoded features onto the UV texture. Combined with other auxiliary inputs, these features allow InstantAlbedo to produce a more accurate, decomposed result.

The image encoder, based on DINO’s ViT 8 × 8 configuration~\cite{caron2021emerging}, combines intermediate features at multiple scales into a unified feature map through upscaling convolutional networks, similar to the cascade architecture in ~\cite{prafull2023materialistic}. We initialize the encoder with weights trained on a dataset with semantic material maps, making it suitable for recognizing materials. Integrated into the InstantAlbedo pipeline, the encoder is fine-tuned jointly with the rest of the model. As shown in \figref{fig:visual_abl}, This design significantly enhances the accuracy of albedo decomposition, even when strong lighting and materials are tightly coupled in the input.

\section{Training}

To train GeoRM, GlossyRM, and InstantAlbedo, we synthesize two separate datasets: one for GeoRM and GlossyRM, and another for InstantAlbedo.

\noindent\textbf{GeoRM and GlossyRM} are trained on a 150K subset of the Objaverse dataset~\cite{deitke2023objaverse}. For each object, we render it from 32 random views, generating measurements, depth, normal, diffuse albedo, roughness, metalness, and mask maps for supervision. We start by training GeoRM, with a two-stage strategy similar to \cite{xu2024instantmesh} (see supplementary material for more details). GeoRM is trained based on differentiable rendering, exclusively with geometry-related losses:
\begin{align}
    \mathcal{L}_{\rm geo}=\frac{1}{N}\sum_{i}&\lambda_z|z_i^{\rm gt}-\hat{z}_i|+\lambda_M\mathcal{L}_{\rm mse}(M_i^{\rm gt}, \hat{M}_i)\notag\\
    &+\lambda_{\mathbf{n}}\mathcal{L}_{\rm lpips}(\mathbf{n}_i^{\rm gt}, \hat{\mathbf{n}}_i),
    \label{eq:geoloss}
\end{align}
where $z_i^{\rm gt}$, $M_i^{\rm gt}$, and $\mathbf{n}_i^{\rm gt}$ are the ground-truth depth, mask, and surface normal for the $i$-th view, and $\hat{z}_i$, $\hat{M}_i$, and $\hat{\mathbf{n}}_i$ are their counterparts rendered from predicted mesh. We randomly select $N$ views from the dataset, with 6 used as input. Next, we train GlossyRM with GeoRM fixed to provide a proxy shape, using the following loss:
\begin{align}
    &\mathcal{L}_{\rm glossy}=\frac{1}{N}\sum_{i}\mathcal{L}_{\rm 0}(\rho_i^{\rm gt}, \hat{\rho}_i)+\mathcal{L}_{\rm 0}(m_i^{\rm gt}, \hat{m}_i),\label{eq:glossyloss}\\
     &\mathcal{L}_{\rm 0}(x,y)=\lambda_1\mathcal{L}_{\rm mse}(x,y)+\lambda_2\mathcal{L}_{\rm lpips}(x, y)+\lambda_3\mathcal{L}_{\rm ssim}(x, y).\notag
\end{align}
Here, $\rho_i^{\rm gt}$ and $m_i^{\rm gt}$ represent the ground-truth roughness and metalness for the $i$-th view, and $\hat{\rho}_i$ and $\hat{m}_i$ are the predicted roughness and metalness rendered from the predicted mesh.

\noindent\textbf{InstantAlbedo} is also trained after GeoRM, but independently from GlossyRM, allowing both models to be trained in parallel. Once GeoRM is trained, we use it to generate 55K shapes from the 150K subset mentioned above, each pre-unwrapped into atlas charts. During training, InstantAlbedo takes the unwrapped mesh and corresponding input images as input, and is trained to predict both the lighting-baked color and the decomposed diffuse albedo. The overall loss is defined as:
\begin{align}
    &\mathcal{L}_{\rm albedo}=\frac{1}{N}\sum_{i}\mathcal{L}_{\rm 0}(\mathbf{c}_i^{\rm gt}, \hat{\mathbf{c}}_i)+\mathcal{L}_{\rm 0}(\mathbf{c_d}_i^{\rm gt}, \hat{\mathbf{c}_d}_i),
    \label{eq:albedoloss}
\end{align}
where $\mathbf{c}_i^{\rm gt}$ and $\mathbf{c_d}_i^{\rm gt}$ denote the ground-truth lighting-baked color and decomposed diffuse albedo for the $i$-th view, and $\hat{\mathbf{c}}_i$ and $\hat{\mathbf{c_d}}_i$ are their predicted counterparts.

\section{Experiments}

\subsection{Experimental settings} We trained ARM on 8 H100 GPUs with a batch size of 1 per GPU over approximately 5 days: two days for GeoRM, two for GlossyRM, and one for InstantAlbedo. We used the Adam optimizer with a learning rate of $4 \times 10^{-5}$ for GeoRM and GlossyRM, and $4 \times 10^{-4}$ for InstantAlbedo. The loss-balancing coefficients were set as $\lambda_z = 0.5$, $\lambda_M = 1.0$, $\lambda_{\mathbf{n}} = 0.2$, $\lambda_1 = 0.7$, $\lambda_2 = 0.3$, and $\lambda_3 = 0.1$. We set $N = 10$ when selecting views.

\subsection{Evaluation settings} We assess the methods on three datasets: GSO~\cite{downs2022google}, OmniObject3D~\cite{wu2023omniobject3d}, and a custom dataset specifically for relightable appearance evaluation. All datasets consist of 3D objects that were unseen during training. For the GSO dataset, we evaluate using all 1,030 available 3D shapes. From the OmniObject3D dataset, we randomly sample up to five shapes from each category, totaling 1,038 shapes for evaluation. The third dataset, introduced for relightable appearance evaluation, includes 100 objects with PBR materials, similar to \cite{zeng2024dilightnet}. For each object, we generate 144 images under varied lighting conditions by rendering 24 random views across six different environmental lightings. To evaluate 3D reconstruction quality, we use both F-score (with a
threshold of 0.1) and Chamfer Distance (CD), comparing the predicted meshes to ground truth meshes. For 2D appearance evaluation, we compute PSNR, SSIM, and LPIPS on rendered images. Since coordinate frames may differ across methods, we align each method’s predicted mesh to the ground truth before calculating metrics. Further alignment details are provided in the supplementary materials.

\subsection{Comparisons}
We first compare our method quantitatively to LGM~\cite{tang2024lgm}, CRM~\cite{wang2024crm}, InstantMesh~\cite{xu2024instantmesh}, MeshFormer~\cite{liu2024meshformer}, and SF3D~\cite{boss2024sf3d} in \tabref{tab:gso_and_omni}. For a fair comparison, all methods are evaluated in a unified single-view to 3D setup. We use the first thumbnail image as input for the GSO dataset, while for the OmniObject3D dataset, we use a rendered image from a random view. For InstantMesh, MeshFormer, and our method, we employ the same Zero123++~\cite{shi2023zero123++} model to generate multi-view images from the single-view input. Since most methods only produce appearance with baked-in lighting, we compare based on lighting-baked color. For SF3D, which lacks lighting-baked vertex color, we render it under even environmental lighting. Our method outperforms others in both geometry and appearance accuracy.

\begin{figure*}
    \centering
    \includegraphics[width=\textwidth]{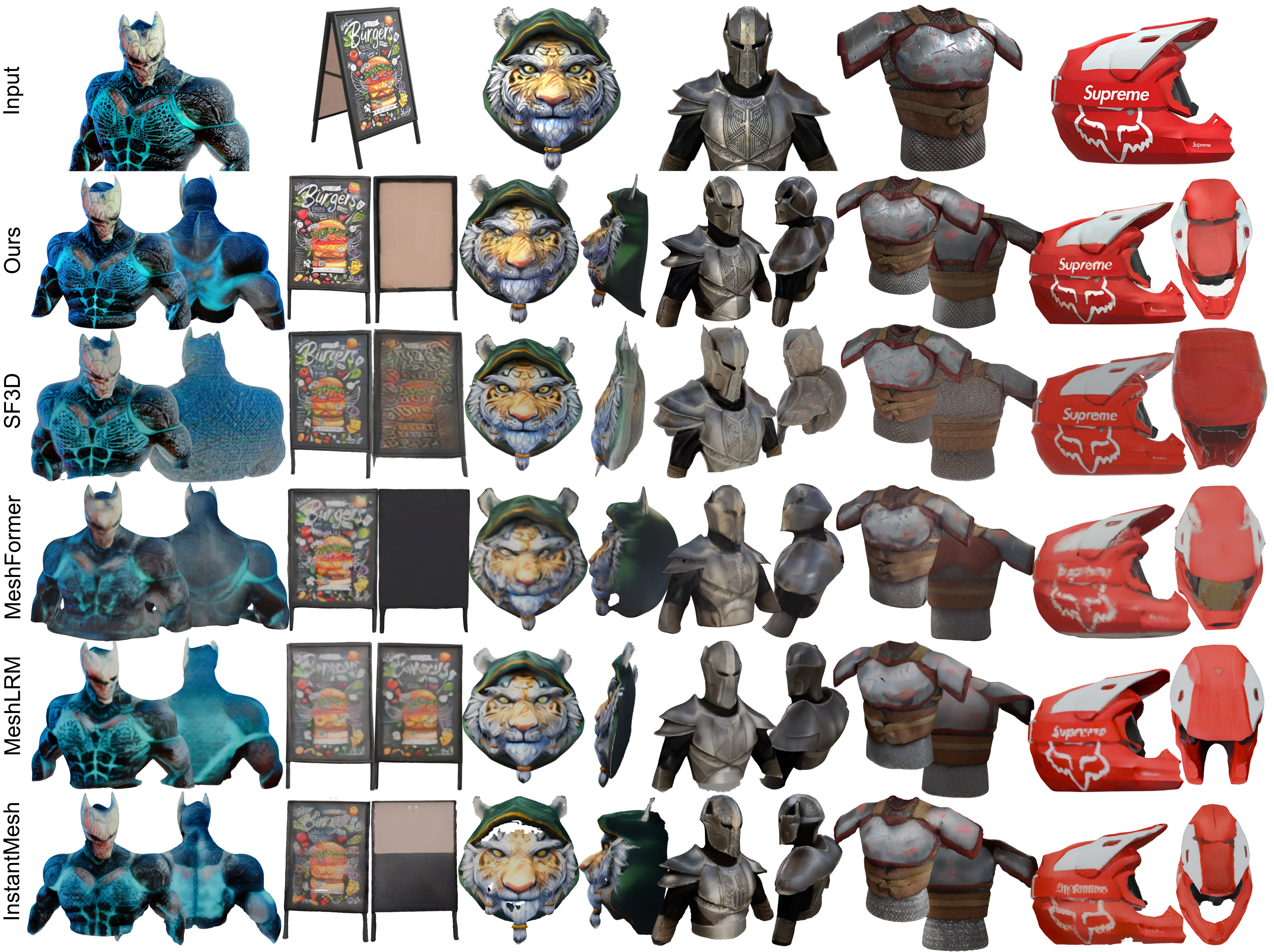}
    \caption{\textbf{Qualitative comparison}. We present examples of single-image 3D generation across different methods. While other methods exhibit blurriness, ARM reconstructs complex patterns with sharp details. Please zoom in to examine the texture quality. Full results, including comparisons with LGM~\cite{tang2024lgm} and CRM~\cite{wang2024crm}, are provided in the supplementary material.}
    \label{fig:color_visual_result}
\end{figure*}

In \figref{fig:color_visual_result}, we provide qualitative examples to visually demonstrate ARM’s superior performance over existing methods. For full results including CRM and LGM, please see our supplementary material. The reconstructed textures from ARM contain significantly richer details, owing to our design in UV texture space. While other methods suffer from blurriness, ARM accurately reconstructs complex and sharp patterns. Some methods, such as SF3D, struggle to generate plausible shape and texture in unseen areas due to training on single-view inputs.

\begin{table*}[h]
\centering
\resizebox{\textwidth}{!}{
\begin{tabular}{ccccccccccc}
\hline
\multicolumn{1}{c|}{\multirow{2}*{Method}} & \multicolumn{5}{c|}{GSO~\cite{downs2022google}} & \multicolumn{5}{c}{OmniObject3D~\cite{wu2023omniobject3d}} \\
\multicolumn{1}{c|}{} & F-Score~$\uparrow$ & CD~$\downarrow$ & PSNR~$\uparrow$ & SSIM~$\uparrow$ & \multicolumn{1}{c|}{LPIPS~$\downarrow$} & F-Score~$\uparrow$ & CD~$\downarrow$ & PSNR~$\uparrow$ &   SSIM~$\uparrow$ &LPIPS~$\downarrow$ \\
\hline
\multicolumn{1}{c|}{LGM~\cite{tang2024lgm}} & 0.784& 0.132  & 18.173 & 0.848 & \multicolumn{1}{c|}{0.207} & 0.801 & 0.127 & 17.979 & 0.843 & 0.229\\
\multicolumn{1}{c|}{CRM~\cite{wang2024crm}}  & 0.893 & 0.091& 19.390 & 0.857 & \multicolumn{1}{c|}{0.180} & 0.845 & 0.110 & 19.083 & \textbf{0.852} & 0.200 \\
\multicolumn{1}{c|}{InstantMesh~\cite{xu2024instantmesh}} & 0.938& 0.065  & 19.744 & 0.858 & \multicolumn{1}{c|}{0.146} & 0.877 & 0.094 & 19.193 & 0.840 & 0.187 \\
\multicolumn{1}{c|}{SF3D~\cite{boss2024sf3d}} & 0.888 & 0.089 & 18.540 & 0.848 & \multicolumn{1}{c|}{0.175} & 0.857 & 0.105 & 18.529 & 0.839 & 0.195 \\
\multicolumn{1}{c|}{MeshFormer~\cite{liu2024meshformer}} & \uline{0.966} & \uline{0.052} & \uline{20.500} & \uline{0.867} & \multicolumn{1}{c|}{\uline{0.141}} & \uline{0.927} & \uline{0.072} & \uline{19.402} & 0.839 & \uline{0.183}\\
\multicolumn{1}{c|}{Ours}  & \textbf{0.968} & \textbf{0.049}& \textbf{21.692} & \textbf{0.880} & \multicolumn{1}{c|}{\textbf{0.137}} & \textbf{0.936} & \textbf{0.067}& \textbf{20.874} & \uline{0.850} & \textbf{0.165} \\
[1pt]\hline

\end{tabular}
}
\caption{\textbf{Quantitative results of single image to 3D}. We evaluate on the GSO~\cite{downs2022google} (1,030 shapes) and OmniObject3D~\cite{wu2023omniobject3d} (1,038 shapes) datasets, reporting results across various metrics. CD denotes Chamfer Distance.}
\label{tab:gso_and_omni}
\end{table*}

\begin{figure*}
    \centering
    \includegraphics[width=\textwidth]{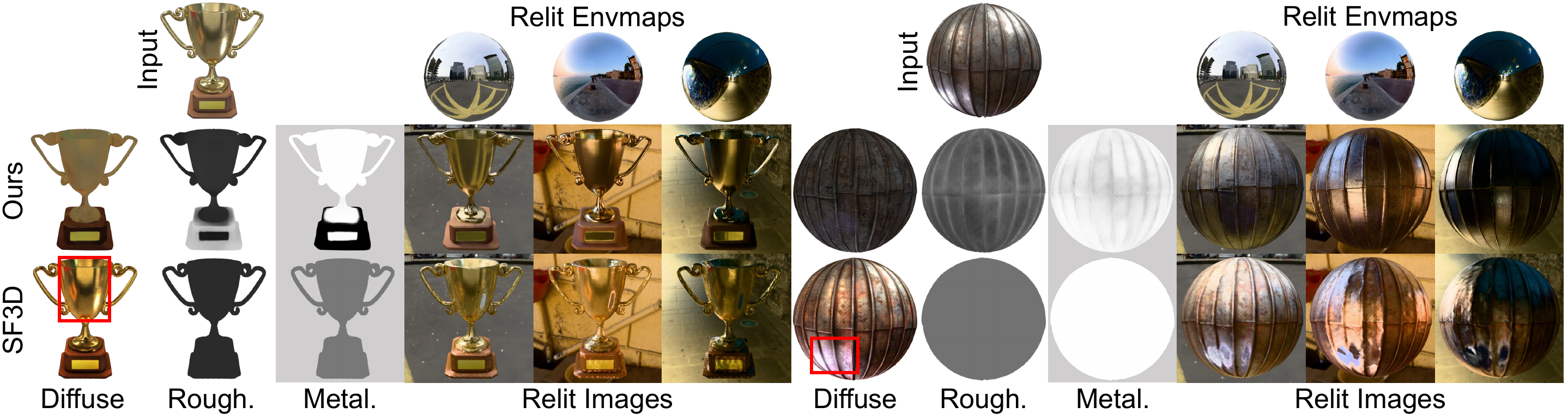}
    \caption{\textbf{PBR comparison}. We compare reconstructed PBR maps and relit images under novel lighting to SF3D~\cite{boss2024sf3d}. While SF3D produces constant roughness and material with lighting baked into the diffuse color (highlighted in the figure), our method generates spatially-varying appearance, with well-separated illumination and materials. See supplementary material for full results.}
    \label{fig:pbr_visual_result}
    \vspace{10pt}
\end{figure*}

\begin{table*}[ht]
    \centering
    \begin{minipage}{\textwidth}
    \begin{minipage}[t]{0.3\textwidth}
        \centering
        \begin{tabular}[t]{ccc}
        \hline
        \multicolumn{1}{c|}{Setting} & PSNR-A~$\uparrow$ & LPIPS-A~$\downarrow$ \\
        \hline
        \multicolumn{1}{c|}{SF3D} & 18.592 & 0.200\\
        \multicolumn{1}{c|}{Ours} & \textbf{21.750} & \textbf{0.171} \\
        \hline
        \end{tabular}
        \caption{\textbf{Quantitative results of relighted appearance.} We evaluate relighted renderings under novel lighting conditions. -A denotes appearance.
        }
        \label{tab:comp_sf3d}
    \end{minipage}
    \hfill
    \begin{minipage}[t]{0.68\textwidth}
        \centering
        \begin{tabular}[t]{ccccc}
        \hline
        \multicolumn{1}{c|}{Setting} & PSNR-A~$\uparrow$ & LPIPS-A~$\downarrow$ & PSNR-D~$\uparrow$ & LPIPS-D~$\downarrow$ \\
        \hline
        \multicolumn{1}{c|}{wo/ Measurements} & 24.780 & 0.104 & 23.398 & 0.114 \\
        \multicolumn{1}{c|}{wo/ Material Prior} & 24.471 & 0.108 & 22.687 & 0.121 \\
        \multicolumn{1}{c|}{wo/ FFC-Net} & 24.612 & 0.110 & 23.360 & 0.123 \\
        \multicolumn{1}{c|}{Baseline} & \textbf{25.074} & \textbf{0.096} & \textbf{24.116} & \textbf{0.098} \\
        \hline
        
        \end{tabular}
        \caption{\textbf{Ablation study.} We evaluate three alternative setups against our full method. -A indicates relighted appearance under novel lighting, and -D denotes predicted diffuse albedo.}
        \label{tab:ablation}
    \end{minipage}
    \end{minipage}
\end{table*}

\begin{figure}
    \centering
    \begin{minipage}[t]{\linewidth}
        \centering
        \begin{minipage}[t]{.25\linewidth}
            \centering
            G.T.
        \end{minipage}%
        \begin{minipage}[t]{.25\linewidth}
            \centering
            Unseen Area
        \end{minipage}%
        \begin{minipage}[t]{.25\linewidth}
            \centering
            w/o FFC-Net
        \end{minipage}%
        \begin{minipage}[t]{.25\linewidth}
            \centering
            w/ FFC-Net
        \end{minipage}%
    \end{minipage}
    \begin{minipage}[t]{\linewidth}
        \centering
        \begin{minipage}[t]{.25\linewidth}
            \centering
            \includegraphics[width=.8\linewidth]{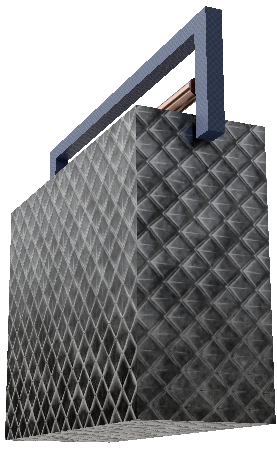}
        \end{minipage}%
        \begin{minipage}[t]{.25\linewidth}
            \centering
            \includegraphics[width=.8\linewidth]{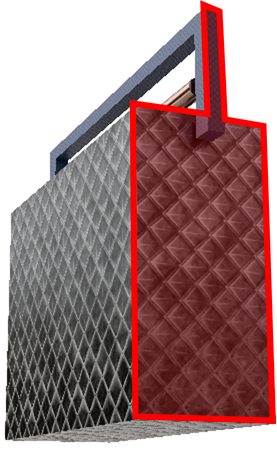}
        \end{minipage}%
        \begin{minipage}[t]{.25\linewidth}
            \centering
            \includegraphics[width=.8\linewidth]{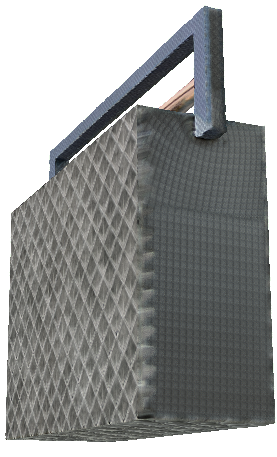}
        \end{minipage}%
        \begin{minipage}[t]{.25\linewidth}
            \centering
            \includegraphics[width=.8\linewidth]{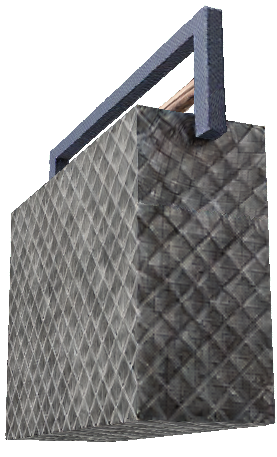}
        \end{minipage}%
    \end{minipage}
    \begin{minipage}[t]{\linewidth}
        \centering
        \begin{minipage}[t]{.25\linewidth}
            \centering
            Input Image
        \end{minipage}%
        \begin{minipage}[t]{.25\linewidth}
            \centering
            G.T. Diffuse
        \end{minipage}%
        \begin{minipage}[t]{.25\linewidth}
            \centering
            w/o Mat. Prior
        \end{minipage}%
        \begin{minipage}[t]{.25\linewidth}
            \centering
            w/ Mat. Prior
        \end{minipage}%
    \end{minipage}
    \begin{minipage}[t]{\linewidth}
        \centering
        \begin{minipage}[t]{.25\linewidth}
            \centering
            \includegraphics[width=.45\linewidth]{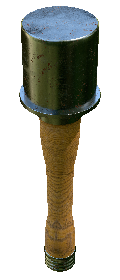}
        \end{minipage}%
        \begin{minipage}[t]{.25\linewidth}
            \centering
            \includegraphics[width=.45\linewidth]{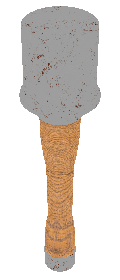}
        \end{minipage}%
        \begin{minipage}[t]{.25\linewidth}
            \centering
            \includegraphics[width=.45\linewidth]{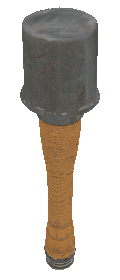}
        \end{minipage}%
        \begin{minipage}[t]{.25\linewidth}
            \centering
            \includegraphics[width=.45\linewidth]{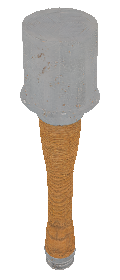}
        \end{minipage}%
    \end{minipage}
    \caption{\textbf{Ablation results for FFC-Net and material prior.} FFC-Net aids in inpainting unseen regions (\textbf{top}), while the material prior improves the diffuse decomposition (\textbf{bottom}), both for the InstantAlbedo stage.}
    \label{fig:visual_abl}
    \vspace{-10pt}
\end{figure}

In \figref{fig:pbr_visual_result}, we compare our reconstructed PBR maps and their relighted images under novel lighting conditions to those produced by SF3D~\cite{boss2024sf3d}, which also reconstructs PBR from single-view input. Our method outperforms SF3D in two key areas: First, when multiple materials are present in the input image, our method reconstructs spatially-varying roughness and metalness, while SF3D generates only constant values, resulting in a homogeneous appearance. Second, SF3D struggles with separating illumination from material properties in the input, leading to baked-in lighting effects. In the cup and ball example, lighting artifacts are embedded in SF3D's reconstructed diffuse albedo, resulting in inaccurate relighting under novel conditions. In contrast, our method successfully decomposes illumination and material, yielding realistic results.

Finally, we present a quantitative comparison of our method and SF3D on the synthetic relighting dataset in \tabref{tab:comp_sf3d}. Our method surpasses SF3D in relit rendering, demonstrating superior accuracy and robustness.

\subsection{Ablations}

We perform both visual and quantitative ablation experiments on the synthetic dataset to evaluate the impact of components in our method. To exclude the effect of multi-view diffusion model, we use ground-truth multi-view images as input. The setups are as follows:

\noindent\textit{w/o measurements}~~~Back-projecting measurements directly visible in the input image provides crucial information for reconstructing textures. In this setup, we omit back-projected image measurements on texture maps.

\noindent\textit{w/o material prior}~~~Incorporating material-aware image encoding adds semantic appearance information, helping to disentangle illumination from materials and improve the accuracy and robustness of appearance decomposition. In this setup, we exclude material features from back-projection.

\noindent\textit{w/o FFC-Net}~~~Since the multi-view images do not cover the full object surface, it is essential to inpaint unseen regions by extracting information from other regions. In this setup, we replace the FFC-Net with a U-Net that has a local receptive field, in contrast to the global scope of the FFC-Net.

As shown in \tabref{tab:ablation}, both modifications degrade quality. Removing the material prior leads to a significant drop in the metrics for decomposed diffuse albedo, while removing the FFC-Net introduces artifacts in unseen areas, reducing perceptual quality notably. \figref{fig:visual_abl} further illustrates the visual impact of removing the material prior and FFC-Net, underscoring that the material prior and FFC-Net play crucial roles in disentangling illumination and material, as well as in inpainting, respectively.

\section{Conclusion and limitation}

We present ARM, a novel method for reconstructing high-quality 3D meshes and PBR maps from sparse-view images, leveraging the advantages of operating in UV space. ARM generates detailed meshes with high-quality textures and spatially-varying materials, outperforming existing methods both qualitatively and quantitatively. However, challenges remain, primarily due to inconsistencies in multi-view images generated by upstream models, which can introduce artifacts in the reconstructed textures. Developing strategies to resolve conflicts across inconsistent views, such as weighting input views based on user specification, is a valuable direction for future exploration.

{
    \small
    \bibliographystyle{ieeenat_fullname}
    \bibliography{main}
}

\clearpage

\maketitlesupplementary

\section{Detailed explanation of \eqnref{eq:brdf}}

ARM models the appearance of object by a spatially varying BRDF described in \eqnref{eq:brdf}. For the microfacet normal distribution term $D$, we use isotropic GGX distribution~\cite{walter2007microfacet}:
$$
D(\mathbf{n},\mathbf{h},\alpha)=\frac{\alpha^2}{\pi((\mathbf{n}\cdot\mathbf{h})(\alpha^2-1)+1)^2},\alpha=\rho^2,
$$
where $\mathbf{n}$ is the half-way vector.  The Geometry function $G$ is based on the Schlick-GGX Geometry function:
$$
G(\mathbf{n}, \mathbf{l}, \mathbf{v}, k)=G_{\rm sub}(\mathbf{n}, \mathbf{l}, k)G_{\rm sub}(\mathbf{n}, \mathbf{v}, k),
$$
where
$$
G_{\rm sub}(\mathbf{n}, \mathbf{v}, k)=\frac{\mathbf{n}\cdot\mathbf{v}}{(\mathbf{n}\cdot\mathbf{v})(1-k)+k}.
$$
Here, $k=(\rho^2+1)^2/8$. Last, the Fresnel term $F$ is
$$
F(\mathbf{v},\mathbf{h})=F_0+(1-F_0)(1-(\mathbf{h}\cdot\mathbf{v}))^5,
$$
where
$$
F_0=m\mathbf{c}_d+(1-m)\mathbf{0.04}.
$$

\section{Details on GeoRM and GlossyRM}

\begin{wrapfigure}[12]{l}{0.45\linewidth}
\vspace{-7pt}
\includegraphics[width=\linewidth]{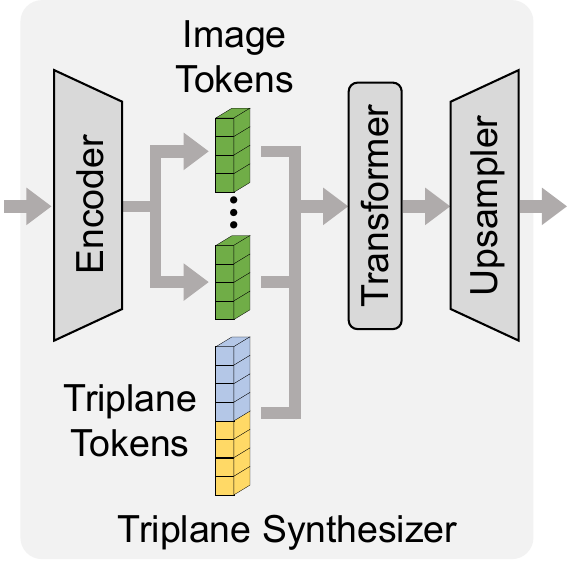}
\vspace{-16pt}
\caption{\small \textbf{Architecture of triplane synthesizer}.}
\label{fig:triplane_synthesizer}
\end{wrapfigure}

GeoRM and GlossyRM are built on the LRM framework, with a super-resolution upsampler added to the triplane synthesizer, as shown in \figref{fig:triplane_synthesizer}.
A pretrained ViT image encoder~\cite{caron2021emerging} converts multi-view input images into image tokens. To make the network aware of camera pose, we add AdaLN camera pose modulation layers to the ViT encoder, following Instant3D~\cite{li2023instant3d}, enabling pose-aware output tokens. The image encoder is jointly fine-tuned during training. The super-resolution upsampler is based on SRResNet~\cite{ledig2017photo}, using four Residual-in-Residual Dense Blocks with a filter size of 512. After these blocks, the upsampling steps consist of three convolutional layers, raising the triplane resolution to 256. Details of the remaining model components, including the encoder and transformer, are provided in \tabref{tab:details_lrm}.

\begin{table}[h]
\renewcommand{\arraystretch}{1.2}
\setlength{\tabcolsep}{10pt}
\centering
\begin{tabular}{lp{0.3\linewidth}}
\hline
Input Views & 6 \\
\hline
Encoder Dim. & 768 \\
\hline
Transformer Dim. & 1024\\
\hline
Transformer Layers & 16\\
\hline
Transformer Heads & 16\\
\hline
Triplane Resolution (Coarse) & 32\\
\hline
Triplane Resolution (Fine) & 256\\
\hline
MLP Hidden Layers & 4\\
\hline
MLP Hidden Dim. & 32\\
\hline
\end{tabular}
\caption{\textbf{Specifications of GeoRM and GlossyRM.} Parameters for each component of the large reconstruction models used in our approach are listed.}
\label{tab:details_lrm}
\end{table}

\begin{figure}
    \centering
    \begin{minipage}[t]{\linewidth}
        \centering
        \begin{minipage}{0.08in}	
            \centering
            \rotatebox{90}{\scalebox{.9}{\small Input}}
        \end{minipage}
        \begin{minipage}[t]{.95\linewidth}
            \centering
            \begin{minipage}{.25\linewidth}	
                \centering
                \includegraphics[width=\linewidth]{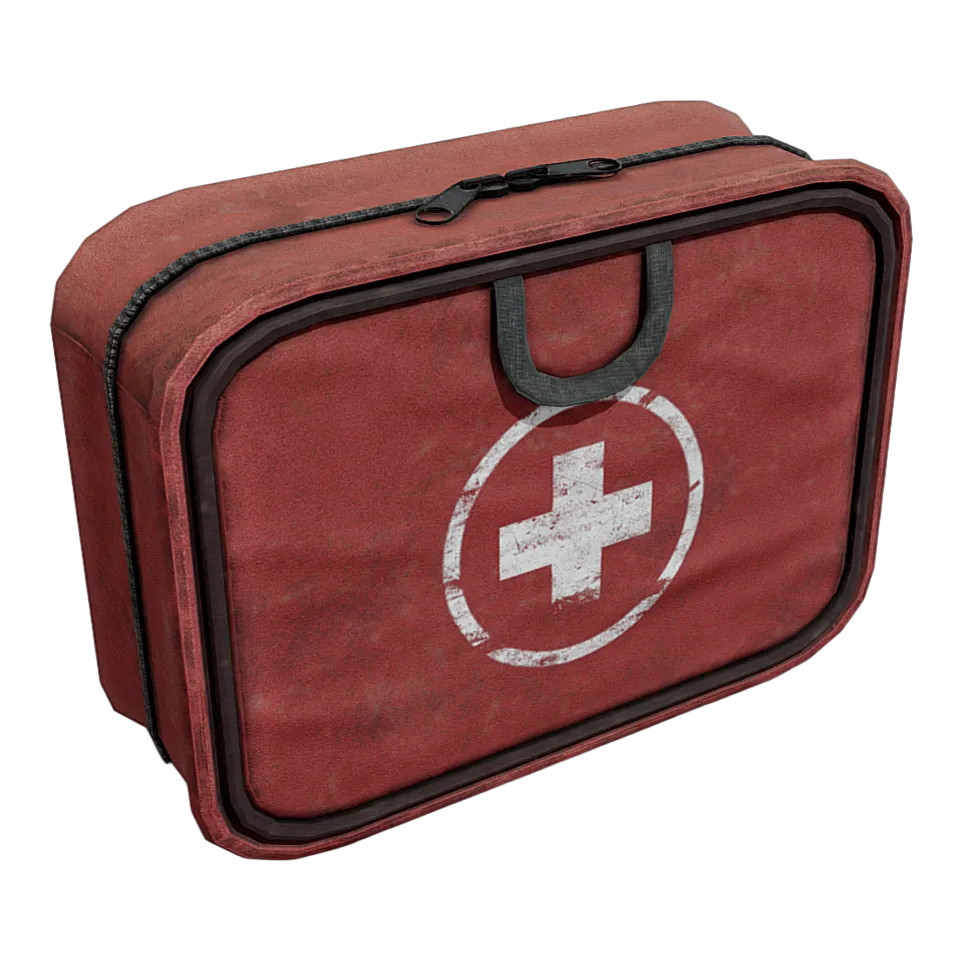}
            \end{minipage}%
            \begin{minipage}{.25\linewidth}	
                \centering
                \includegraphics[width=\linewidth]{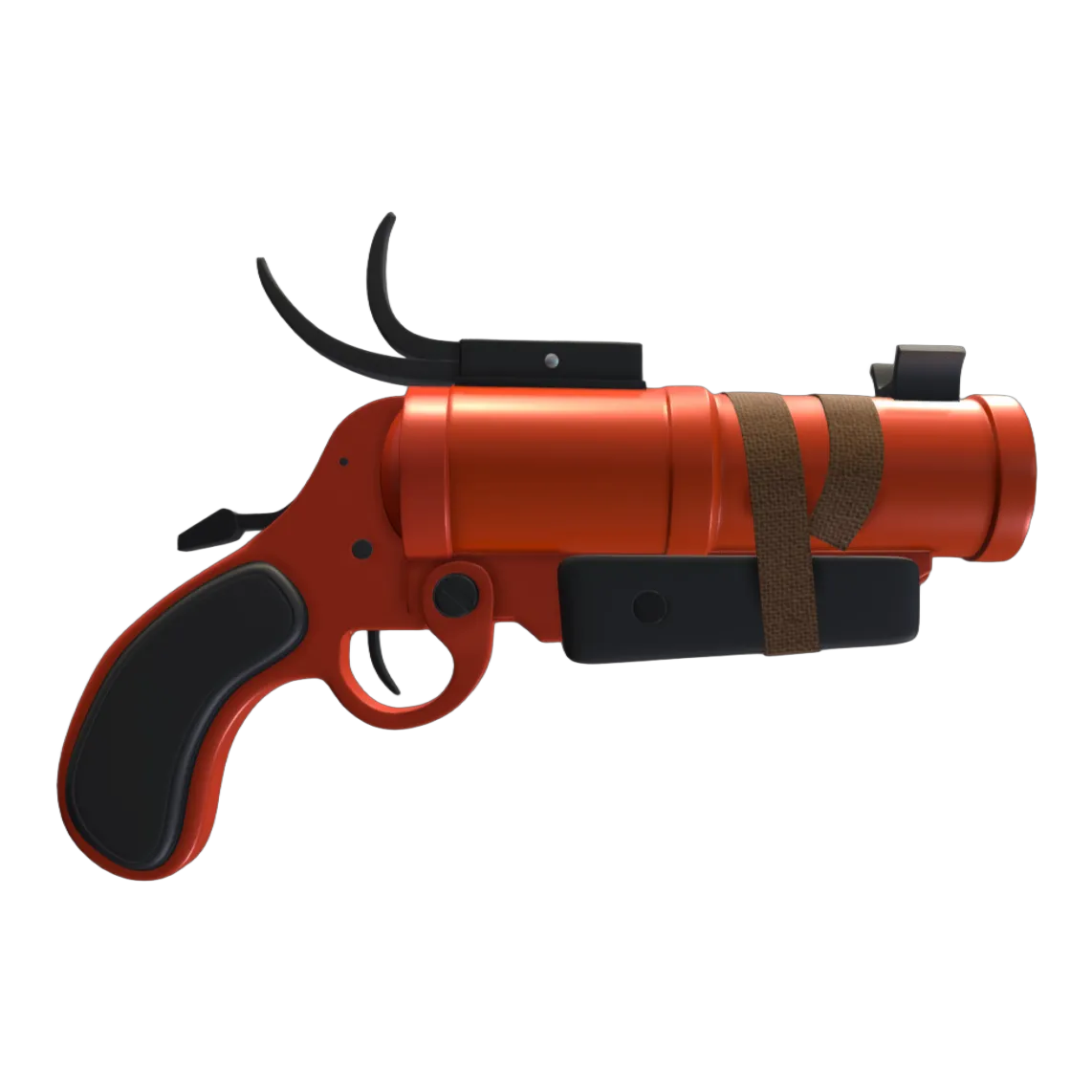}
            \end{minipage}%
            \begin{minipage}{.25\linewidth}	
                \centering
                \includegraphics[width=\linewidth]{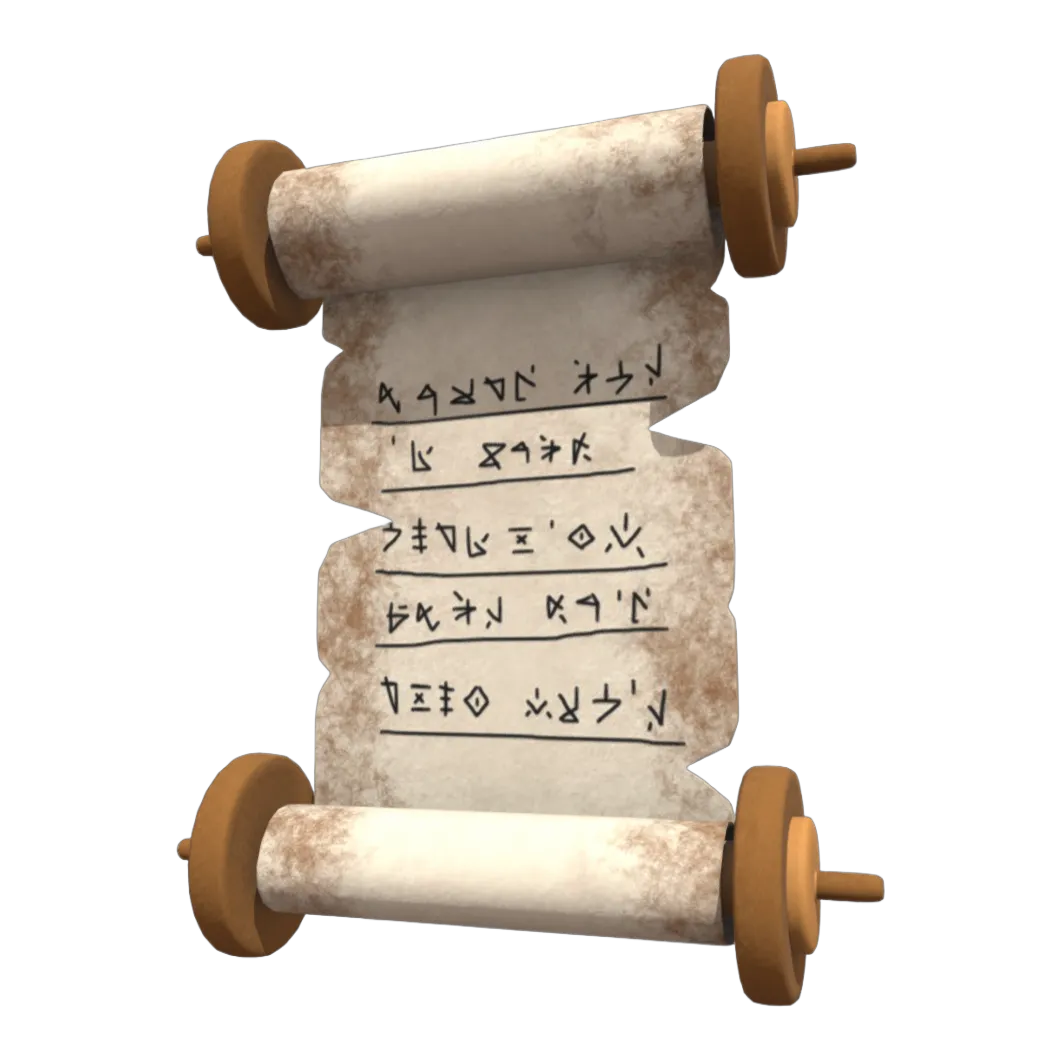}
            \end{minipage}%
            \begin{minipage}{.25\linewidth}	
                \centering
                \includegraphics[width=\linewidth]{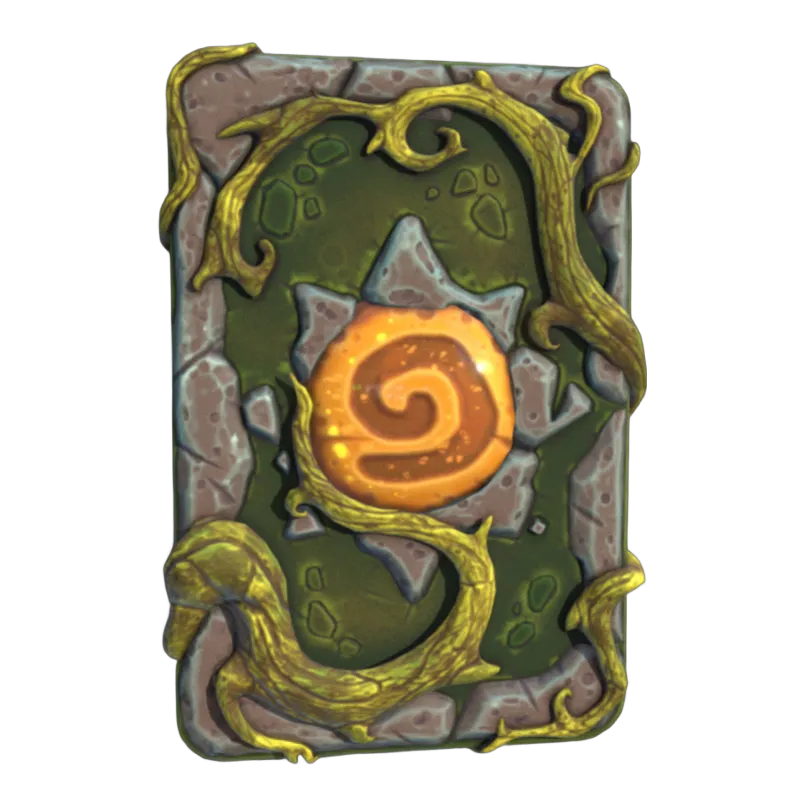}
            \end{minipage}%
        \end{minipage}
    \end{minipage}
    \begin{minipage}[t]{\linewidth}
        \centering
        \begin{minipage}{0.08in}	
            \centering
            \rotatebox{90}{\scalebox{.9}{\small InstantAlbedo $\rho$}}
        \end{minipage}
        \begin{minipage}[t]{.95\linewidth}
            \centering
            \begin{minipage}{.25\linewidth}	
                \centering
                \includegraphics[width=\linewidth]{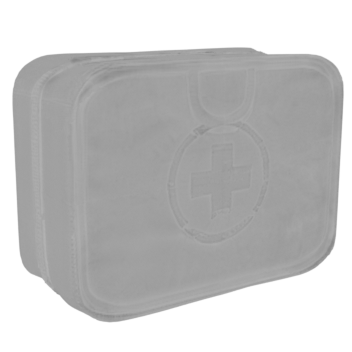}
            \end{minipage}%
            \begin{minipage}{.25\linewidth}	
                \centering
                \includegraphics[width=\linewidth]{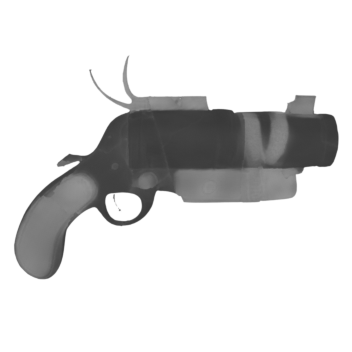}
            \end{minipage}%
            \begin{minipage}{.25\linewidth}	
                \centering
                \includegraphics[width=\linewidth]{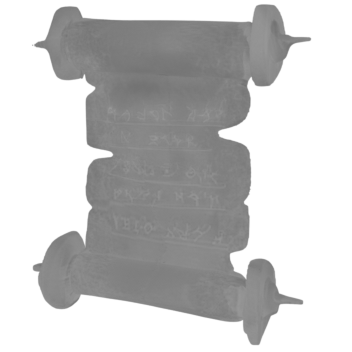}
            \end{minipage}%
            \begin{minipage}{.25\linewidth}	
                \centering
                \includegraphics[width=\linewidth]{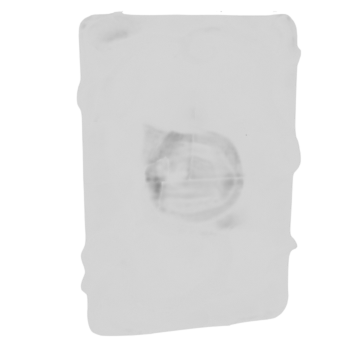}
            \end{minipage}%
        \end{minipage}
    \end{minipage}
    \begin{minipage}[t]{\linewidth}
        \centering
        \begin{minipage}{0.08in}	
            \centering
            \rotatebox{90}{\scalebox{.9}{\small GlossyRM $\rho$}}
        \end{minipage}
        \begin{minipage}[t]{.95\linewidth}
            \centering
            \begin{minipage}{.25\linewidth}	
                \centering
                \includegraphics[width=\linewidth]{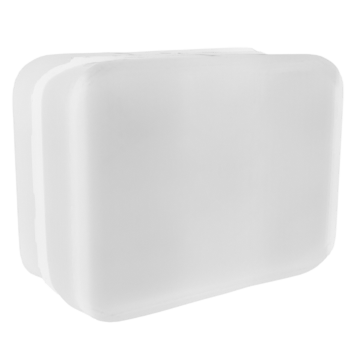}
            \end{minipage}%
            \begin{minipage}{.25\linewidth}	
                \centering
                \includegraphics[width=\linewidth]{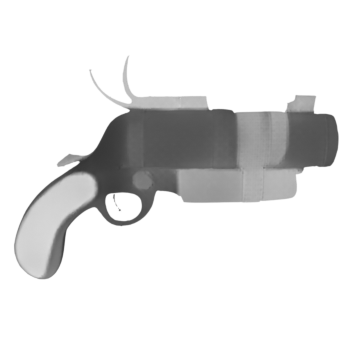}
            \end{minipage}%
            \begin{minipage}{.25\linewidth}	
                \centering
                \includegraphics[width=\linewidth]{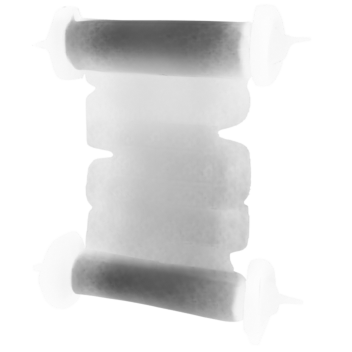}
            \end{minipage}%
            \begin{minipage}{.25\linewidth}	
                \centering
                \includegraphics[width=\linewidth]{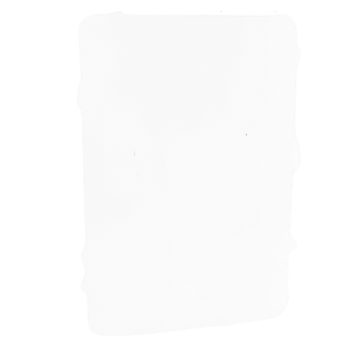}
            \end{minipage}%
        \end{minipage}
    \end{minipage}
    \begin{minipage}[t]{\linewidth}
        \centering
        \begin{minipage}{0.08in}	
            \centering
            \rotatebox{90}{\scalebox{.9}{\small InstantAlbedo $m$}}
        \end{minipage}
        \begin{minipage}[t]{.95\linewidth}
            \centering
            \begin{minipage}{.25\linewidth}	
                \centering
                \includegraphics[width=\linewidth]{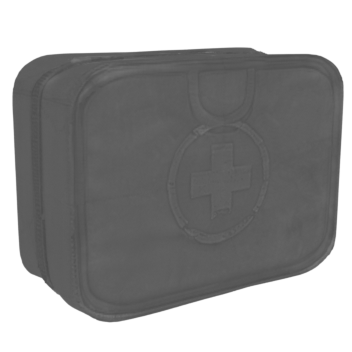}
            \end{minipage}%
            \begin{minipage}{.25\linewidth}	
                \centering
                \includegraphics[width=\linewidth]{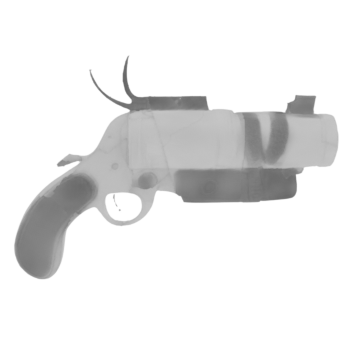}
            \end{minipage}%
            \begin{minipage}{.25\linewidth}	
                \centering
                \includegraphics[width=\linewidth]{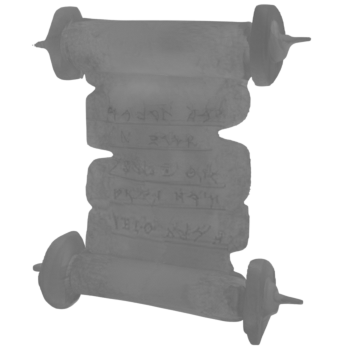}
            \end{minipage}%
            \begin{minipage}{.25\linewidth}	
                \centering
                \includegraphics[width=\linewidth]{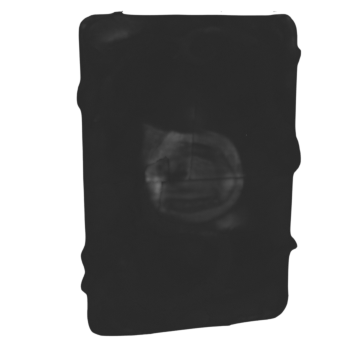}
            \end{minipage}%
        \end{minipage}
    \end{minipage}
    \begin{minipage}[t]{\linewidth}
        \centering
        \begin{minipage}{0.08in}	
            \centering
            \rotatebox{90}{\scalebox{.9}{\small GlossyRM $m$}}
        \end{minipage}
        \begin{minipage}[t]{.95\linewidth}
            \centering
            \begin{minipage}{.25\linewidth}	
                \centering
                \includegraphics[width=\linewidth]{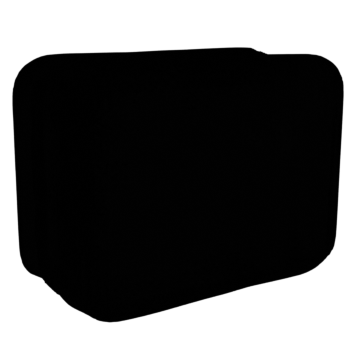}
            \end{minipage}%
            \begin{minipage}{.25\linewidth}	
                \centering
                \includegraphics[width=\linewidth]{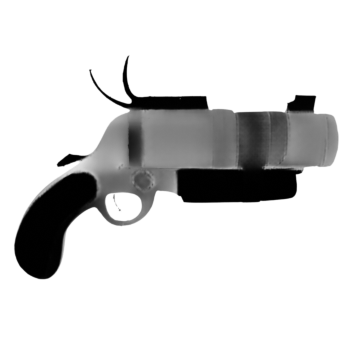}
            \end{minipage}%
            \begin{minipage}{.25\linewidth}	
                \centering
                \includegraphics[width=\linewidth]{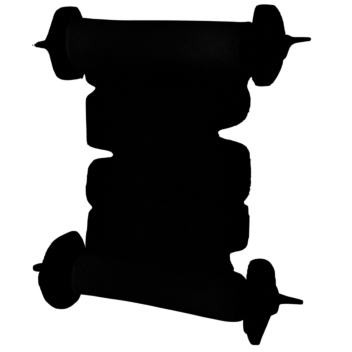}
            \end{minipage}%
            \begin{minipage}{.25\linewidth}	
                \centering
                \includegraphics[width=\linewidth]{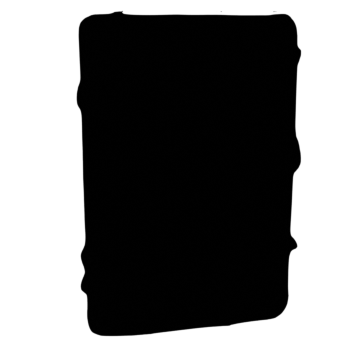}
            \end{minipage}%
        \end{minipage}
    \end{minipage}
    \caption{\textbf{Comparison with unified material prediction.} ARM separates the prediction of roughness and metalness by using GlossyRM, rather than predicting all material parameters within InstantAlbedo. We compare the differences between two approaches. InstantAlbedo tends to predict only intermediate values for roughness and metalness, making it difficult to produce extreme values close to $0$ or $1$, particularly for non-metallic objects.}
    \label{fig:abl_unified}
\end{figure}

\begin{table*}[h]
\centering
\resizebox{\textwidth}{!}{
\begin{tabular}{c ccc ccc ccc}
\hline
\multicolumn{1}{c|}{Method} & PSNR-D~$\uparrow$ & SSIM-D~$\uparrow$ & \multicolumn{1}{c|}{LPIPS-D~$\downarrow$} & PSNR-$\rho$~$\uparrow$ & SSIM-$\rho$~$\uparrow$ & \multicolumn{1}{c|}{LPIPS-$\rho$~$\downarrow$} & PSNR-$m$~$\uparrow$ & SSIM-$m$~$\uparrow$ & LPIPS-$m$~$\downarrow$ \\
\hline
\multicolumn{1}{c|}{SF3D~\cite{boss2024sf3d}} & 16.937 & 0.834 & \multicolumn{1}{c|}{0.205} & 18.012 & 0.873 & \multicolumn{1}{c|}{0.202} & 20.433 & 0.862 & 0.153 \\
\multicolumn{1}{c|}{Ours} & \textbf{21.108} & \textbf{0.844} & \multicolumn{1}{c|}{\textbf{0.178}} & \textbf{19.565} & \textbf{0.883} & \multicolumn{1}{c|}{\textbf{0.165}} & \textbf{21.866} & \textbf{0.883} & \textbf{0.145} \\
[1pt]\hline

\end{tabular}
}
\caption{\textbf{Quantitative Results of Reconstructed PBR Maps.} We report metrics comparing the predicted PBR maps with ground truth. Due to the high ambiguity in appearance decomposition, where multiple valid decompositions can explain the same shaded image, we only provide indicative scores in the supplementary material. Here, -D represents diffuse albedo, -$\rho$ denotes roughness, and -$m$ denotes metalness.}
\label{tab:pbr_quant}
\end{table*}

While GeoRM and GlossyRM share the same architecture, they are trained as two distinct models. For GeoRM, we adopt a two-stage training strategy similar to \cite{xu2024instantmesh}. In the first stage, we load pretrained weights for all components except the newly introduced super-resolution module and train using a volume rendering loss. In the second stage, we employ differentiable marching cubes to extract iso-surface from the queried density grid, followed by rendering with a differentiable rasterizer~\cite{Laine2020diffrast}.

After training GeoRM, we proceed to train GlossyRM while keeping GeoRM fixed. Specifically, we first use GeoRM to generate the 3D shape from the multi-view input. Then, for each vertex on this generated shape, we retrieve features from GlossyRM's triplane and feed them into the decoding MLP to predict roughness and metalness. These per-vertex properties are then used to render multi-view images, with a loss computed against ground-truth images to guide GlossyRM’s training. For faster convergence, GlossyRM is initialized with GeoRM's weights at the start of training.
\begin{figure*}
    \centering
    \includegraphics[width=\textwidth]{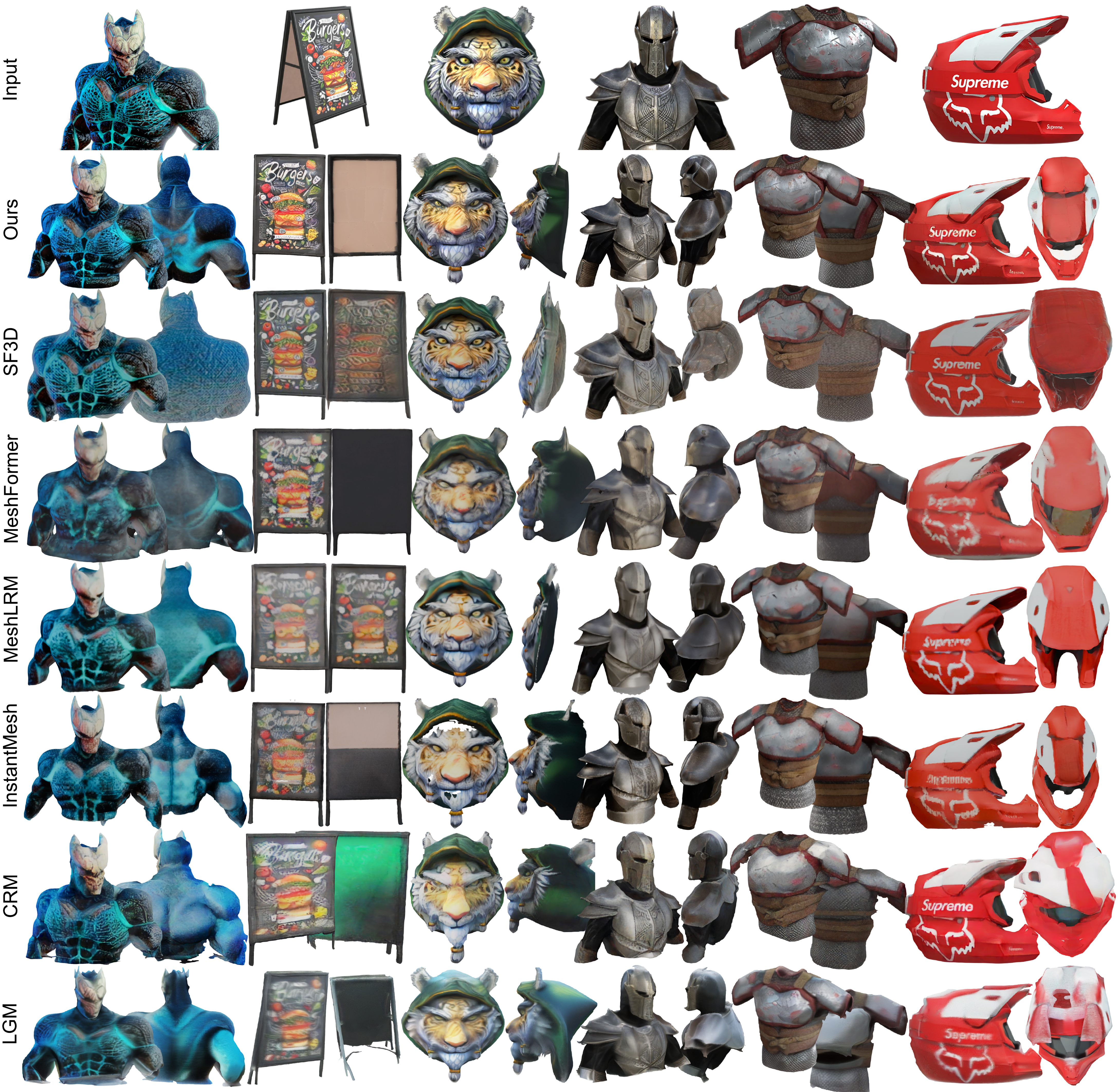}
    \caption{\textbf{Qualitative comparison}. We present examples of single-image 3D generation across different methods. While other methods exhibit blurriness, ARM reconstructs complex patterns with sharp details. Please zoom in to examine the texture quality.}
    \label{fig:color_visual_result_full}
\end{figure*}

\begin{figure*}
    \centering
    \includegraphics[width=\textwidth]{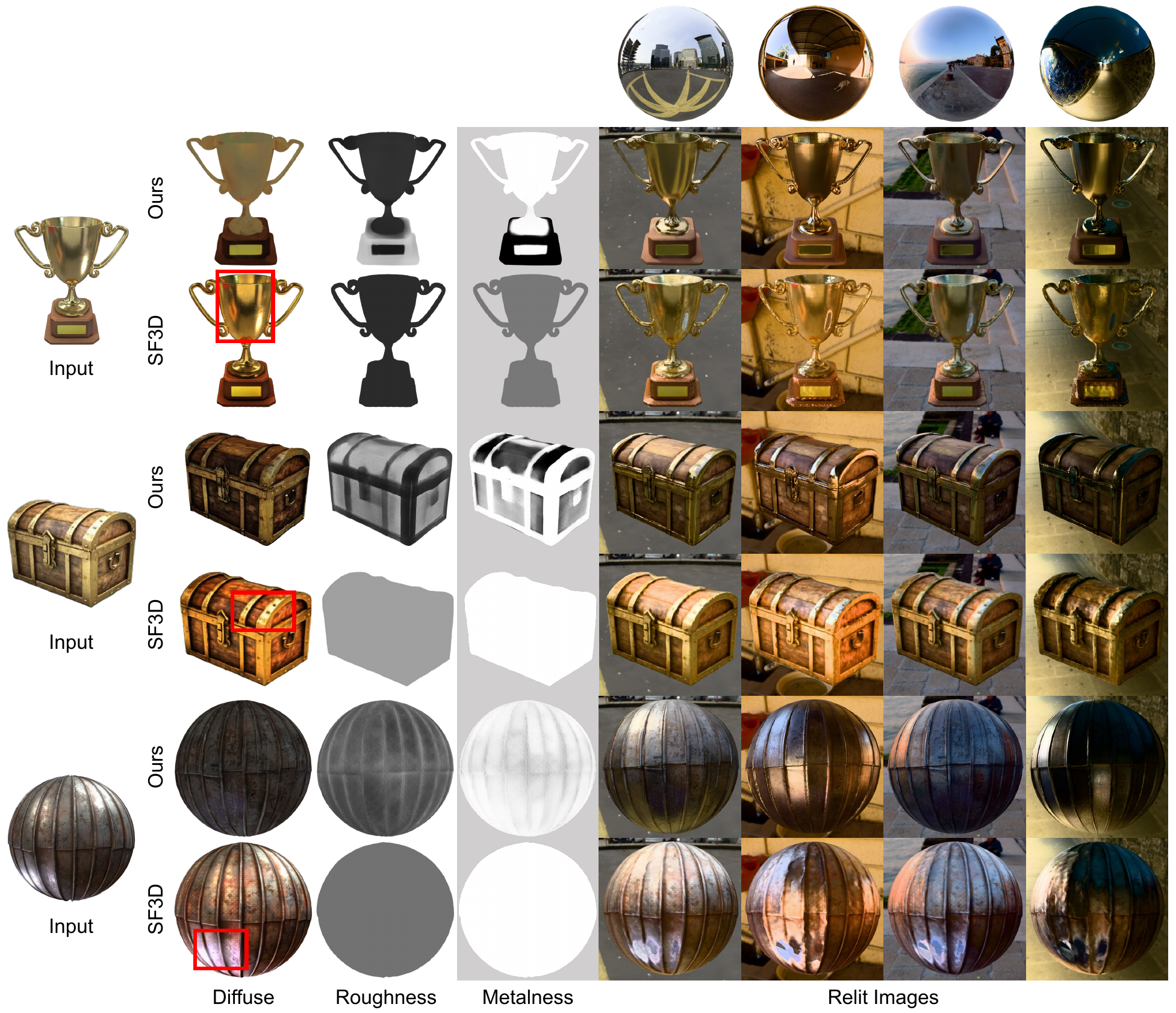}
    \caption{\textbf{PBR comparison}. We compare reconstructed PBR maps and relit images under novel lighting to SF3D~\cite{boss2024sf3d}. While SF3D produces constant roughness and material with lighting baked into the diffuse color (highlighted in the figure), our method generates spatially-varying appearance, with well-separated illumination and materials.}
    \label{fig:pbr_visual_result_full}
\end{figure*}


\section{Unified material prediction}
ARM separates PBR parameter prediction into two networks: InstantAlbedo for diffuse albedo and GlossyRM for roughness and metalness. Although predicting all material properties within InstantAlbedo might seem more straightforward, our experiments indicate that this approach results in inaccurate material decomposition, as shown in \figref{fig:abl_unified}. InstantAlbedo tends to predict only intermediate values for roughness and metalness, making it difficult to produce extreme values close to $0$ or $1$, particularly for non-metallic objects. Notably, for SVBRDF, human perception is generally more sensitive to spatial variations (subtle pixel changes within textures) than to angular variations (subtle changes of lighting and view direction in BRDF). By leveraging GlossyRM, which has ample network capacity, our method effectively produces realistic appearances, with InstantAlbedo capturing the fine-grained details in diffuse albedo.

\section{Details on InstantAlbedo}
The InstantAlbedo framework comprises three main networks: a material-aware image encoder, a U-Net, and an FFC-Net. The material-aware image encoder is based on \cite{prafull2023materialistic}, excluding the user reference injection and cross-attention layers. For the FFC-Net, we use a ResNet-like architecture~\cite{he2016deep} with 3 downsampling blocks, 4 residual blocks, and 3 upsampling blocks. In our model, the residual blocks utilize FFC with a filter size of 512.

\section{Dataset selection}

GeoRM and GlossyRM are trained on a 150K subset of the Objaverse dataset~\cite{deitke2023objaverse}. This subset is carefully curated based on the following criteria to ensure high-quality training data:
\begin{enumerate}
    \item Each selected object must include a roughness map or a metalness map. This requirement ensures that the objects have sufficient material data for training GlossyRM.
    \item The object must not be a point cloud, nor a sparse or small object with low occupancy (fewer than 10 pixels per rendered view).
    \item Low-quality objects, such as scanned indoor data or large scenes with multiple objects, are excluded. 
\end{enumerate}

\section{Shape alignment}

During evaluation, we align each method’s predicted meshes to the ground truth meshes before calculating metrics, as coordinate frames may differ across methods. Following MeshFormer~\cite{liu2024meshformer}, we use a two-step alignment based on the evaluation metric. First, we normalize both ground truth and predicted meshes to fit within a bounding box in the range $[-1, 1]^3$. Then, we uniformly sample rotations in $[0, 2\pi)$ and scales in $[0.7, 1.4]$ for initialization, refining the alignment using the Iterative Closest Point (ICP) algorithm. We select the alignment with the highest evaluation score.

Once aligned, we compute metrics for each method. For 3D metrics, we sample 100,000 points on both the ground truth and predicted meshes to calculate the F-score and Chamfer Distance, setting a threshold of 0.1 for the F-score. To evaluate texture quality, we compute PSNR, SSIM, and LPIPS between images rendered from the reconstructed mesh and ground truth. We sample 32 camera poses in a full 360-degree view around the object, rendering RGB images at a resolution of 320$\times$320. Since we use the VGG model for LPIPS loss during training, we use the Alex model for LPIPS evaluation.

\section{Additional results}

In \tabref{tab:pbr_quant}, We report quantitative metrics comparing the predicted PBR maps with ground truth, using SF3D and our method. Due to the high ambiguity in appearance decomposition, where multiple valid decompositions can explain the same shaded image, we only provide indicative scores in the supplementary material.

\figref{fig:color_visual_result_full} presents complete qualitative examples, including comparisons with LGM and CRM. In \figref{fig:pbr_visual_result_full}, we provide further examples along with additional relighting results.

\end{document}